\documentclass[sigconf]{acmart}

\copyrightyear{2018} 
\acmYear{2018} 
\setcopyright{acmcopyright}
\acmConference[KDD 2018]{24th ACM SIGKDD International Conference on Knowledge Discovery \& Data Mining}{August 19--23, 2018}{London, United Kingdom}
\acmBooktitle{KDD 2018: 24th ACM SIGKDD International Conference on Knowledge Discovery \& Data Mining, August 19--23, 2018, London, United Kingdom}
\acmPrice{15.00}
\acmDOI{10.1145/3219819.3220042}
\acmISBN{978-1-4503-5552-0/18/08} 
\usepackage{booktabs} 

\usepackage{graphicx}
\usepackage{amssymb}
\usepackage{amsthm}
\usepackage{adjustbox}
\usepackage[font=small]{caption}
\usepackage{multirow} 

\usepackage{algorithmic}
\usepackage{algorithm}
\usepackage{balance}

\DeclareMathOperator*{\argmin}{arg\,min}

\begin{document}
\title[Learning Representations of Ultrahigh-dimensional Data]{Learning Representations of Ultrahigh-dimensional Data for Random Distance-based Outlier Detection}

\author{Guansong Pang}
\affiliation{%
  \institution{University of Technology Sydney}
  \city{Sydney}
  \country{Australia}
  \postcode{2007}
}
\email{pangguansong@gmail.com}

\author{Longbing Cao}
\affiliation{%
  \institution{University of Technology Sydney}
  \city{Sydney}
  \country{Australia}
  \postcode{2007}
}
\email{longbing.cao@uts.edu.au}

\author{Ling Chen}
\affiliation{%
  \institution{University of Technology Sydney}
  \city{Sydney}
  \country{Australia}
  \postcode{2007}
}
\email{ling.chen@uts.edu.au}

\author{Huan Liu}
\affiliation{%
  \institution{Arizona State University}
  \city{Tempe}
  \country{United States}
}
\email{huan.liu@asu.edu}







\begin{abstract}

Learning expressive low-dimensional representations of ultrahigh-dimensional data, e.g., data with thousands/millions of features, has been a major way to enable learning methods to address the curse of dimensionality. However, existing \textit{unsupervised} representation learning methods mainly focus on preserving the data regularity information and learning the representations independently of subsequent outlier detection methods, which can result in suboptimal and unstable performance of detecting irregularities (i.e., outliers). 

This paper introduces a ranking model-based framework, called RAMODO, to address this issue. RAMODO unifies representation learning and outlier detection to learn low-dimensional representations that are \textit{tailored} for a state-of-the-art outlier detection approach - the random distance-based approach. This customized learning yields more optimal and stable representations for the targeted outlier detectors. Additionally, RAMODO can leverage little labeled data as \textit{prior knowledge} to learn more expressive and application-relevant representations. We instantiate RAMODO to an efficient method called REPEN to demonstrate the performance of RAMODO. 

Extensive empirical results on eight real-world ultrahigh dimensional data sets show that REPEN (i) enables a random distance-based detector to obtain significantly better AUC performance and two orders of magnitude speedup; (ii) performs substantially better and more stably than four state-of-the-art representation learning methods; and (iii) leverages less than 1\% labeled data to achieve up to 32\% AUC improvement.


\end{abstract}

%
%



\maketitle

\section{Introduction}

Outlier detection, which is referred to as the process of identifying data objects that deviate significantly from the majority of data objects, can offer important insights into broad applications in a range of domains such as medical diagnosis, fraud detection, and information security. Many of these applications often have ultrahigh dimensionality, e.g., thousands of features in detecting
abnormal bioactivities \cite{danziger2007p53mutant}, hundreds of thousands of features in corporate fraud detection \cite{junque2014corporatefraud}, and millions of features in detecting malicious URLs \cite{ma2009url}. Such ultrahigh-dimensional data presents significant challenges to existing outlier detection methods due to the curse of dimensionality \cite{zimek2012survey}. 

One straightforward yet challenging solution is to map such high-dimensional data sets into low-dimensional representations that preserve the relevant information for subsequent learning tasks. Many unsupervised representation learning techniques have been introduced to address this issue, such as spectral-based methods, neural network-based methods, and manifold learning \cite{bengio2013representation,van2009dr}. However, most of these methods focus on preserving the data regularity information (e.g., data reconstruction/proximity information) for learning tasks like clustering and data compression. They may therefore fail to retain the important information for uncovering the irregularities (i.e., outliers). A few studies (e.g., \cite{rahmani2017cop}) have attempted to learn representations for identifying outliers in very recent years. However, they learn the representations independently of subsequent outlier detection methods. As a result, the optimal representations they produce may be suboptimal to a given specific outlier detection method, which leads to ineffective and unstable performance of the outlier detector. 

Also, since these techniques focus on unsupervised learning, it is difficult for them to incorporate application-specific knowledge (e.g., a few labeled outliers) into the representation learning. When such prior knowledge is available as in many real-world outlier detection applications, the valuable information cannot be used by these representation learning techniques.

In an attempt to address the above issues, this paper introduces a RAnking MOdel-based representation learning framework to learn representations of ultrahigh-dimensional data for Distance-based Outlier detection methods (RAMODO). RAMODO incorporates random distance-based outlier detection methods into the objective function of its representation learning to learn customized representations for such outlier scoring methods. \textit{Random distance-based outlier detection} defines the outlierness of a data object based on its distance to the data objects in a random subsample. This approach has shown state-of-the-art accuracy and scalability \cite{sugiyama2013dbsubsample,pang2015lesinn,ting2017defying}, but it still suffers from the curse of dimensionality. RAMODO therefore chooses random distance-based methods as its representation learning target to introduce state-of-the-art distance-based methods for ultrahigh-dimensional outlier detection. Moreover, RAMODO can be easily extended to leverage the application-specific knowledge to learn more expressive and application-relevant representations.

RAMODO is implemented by a method called REPEN that learns customized REPresentations for a random nEarest Neighbor distance-based method, Sp \cite{sugiyama2013dbsubsample}. REPEN defines its objective function using Sp-based outlier scores to guide the representation learning, resulting in low-dimensional representations that are tailored for Sp. While Sp is chosen for its state-of-the-art effectiveness and efficiency \cite{sugiyama2013dbsubsample}, RAMODO can also be customized for other random distance-based methods.

Accordingly, this paper makes three major contributions:

\begin{enumerate}
    \item We introduce the RAMODO framework to learn customized low-dimensional representations of ultrahigh-dimensional data for random distance-based outlier detectors. Unlike existing methods that preserve the regularity information and separate representation learning from subsequent outlier detectors, RAMODO unifies representation learning and outlier detection to learn a small set of features that are tailored for the random distance-based detectors. As a result, RAMODO can learn better representations for the detectors with more effective and stable performance.
    \item The RAMODO framework is instantiated into a method called REPEN to learn customized representations for one state-of-the-art random distance-based method, Sp. Sp has provable error bounds and is highly scalable, which enables REPEN to learn the representations with an upper error bound and scale up to large ultrahigh-dimensional data.
    \item We further introduce a method for REPEN to incorporate a small set of labeled outliers as application-specific knowledge, which helps the REPEN-enabled Sp identify application-relevant outliers, rather than data noises or uninteresting data objects due to the lack of such prior knowledge. This capability results in practical solutions in many real-world applications where a few labeled outliers are available.
\end{enumerate}

Extensive empirical results on eight real-world data sets with thousands to millions of features and two sets of synthetic data show that REPEN (i) enables the original distance-based outlier detector to obtain significantly better AUC performance and two orders of magnitude speedup; (ii) performs substantially better and more stably than four state-of-the-art representation learning competitors; (iii) achieves up to 32\% AUC improvement by leveraging less than 1\% labeled outliers as prior knowledge; (iv) performs stably w.r.t. a wide range of representation dimensions; and (v) obtains linear time complexity w.r.t. both data size and dimensionality.

\section{Related Work}

\subsection{Distance-based Outlier Detection}
Distance-based outlier detection is arguably one of the most widely-used detection approaches \cite{campos2016evaluation}. Some very popular distance-based methods include $K$-th nearest neighbor distance- and average $K$ nearest neighbors distance-based methods \cite{bay2003distance}. This type of methods has time complexity quadratic w.r.t. data size. The time complexity may be reduced to be nearly linear by using indexing \cite{bay2003distance} or distributed computing techniques \cite{cao2017distributeddistance}. Recent studies \cite{sugiyama2013dbsubsample,pang2015lesinn,ting2017defying} show that random distance-based methods or distance-based ensemble methods can achieve not only a similar time complexity reduction but also low false positive errors, resulting in scalable state-of-the-art distance-based detectors. However, these techniques still do not address the curse of dimensionality issue. Subspace-based approaches \cite{aggarwal2005effective,keller2012hics} define outlierness using a set of relevant feature subspaces to avoid the curse of dimensionality. Outlying feature selection, which retains a feature subset that is relevant to outlier detection, emerges as an alternative solution to subspace-based methods \cite{azmandian2012icdm,pang2017hour,pang2016dsfs,pang2018cinfo}. However, both approaches are mainly focused on data sets with tens/hundreds of features due to their prohibitive subspace search in the ultrahigh-dimensional space. 



\subsection{Unsupervised Representation Learning for Outlier Detection} 

Numerous unsupervised representation learning methods have been proposed to learn low-dimensional representations of high-dimensional data \cite{bengio2013representation,van2009dr}. They include: spectral-based methods, like principal  component analysis (PCA) and its variants \cite{candes2011robustpca,jian2017embedding}; neural network-based methods, like autoencoder and its variants \cite{hinton2006ae,makhzani2015ae}; manifold learning, like locally linear embedding (LLE) and Hessian LLE \cite{donoho2003hessian}; random projection, like sparse random projection \cite{li2006srp}, to name a few. However, these methods can be biased by the presence of outliers, since they treat the inliers and outliers equally. Also, these methods are mainly designed to preserve the information of data regularities for unsupervised learning tasks like clustering and data compression. Their resultant representations may therefore ignore important information for uncovering the irregularities. 

In recent years, robust PCA and robust autoencoder methods have been introduced to learn robust representations to reduce the bias caused by the outliers \cite{vincent2010dae,candes2011robustpca}. However, these methods mainly address the bias issue. The very recent coherent pursuit in \cite{rahmani2017cop} and the combination of robust PCA and autoencoders \cite{zhou2017anomaly} help address the above both issues. However, these two methods involve the costly eigen analysis or alternative optimization, making it unscalable to large ultrahigh-dimensional data. In addition, the method in \cite{zhou2017anomaly} needs to tune its parameters in a semi-supervised way. Lastly, all the above methods may produce suboptimal representations for a specific outlier detection method, as they ignore the subsequent outlier detection when learning the data representations.


\subsection{Using Labeled Outliers as Prior Knowledge}

One problem with unsupervised outlier detection methods is that many of the outliers they identify are data noises or uninteresting data objects due to the lack of prior knowledge of irregularities \cite{aggarwal2017supervised}. Building the outlier detectors with application-specific knowledge (e.g., labeled outliers) may help identify application-relevant outliers. Related studies have attempted to incorporate some labeled outliers into graph outlier detection by belief propagation \cite{mcglohon2009snare,tamersoy2014guilt}. Converting ultrahigh-dimensional data into $K$-nearest-neighbor graph may facilitate the adoption of this technique, but such conversion faces big challenges due to the difficulty of obtaining accurate distances in the high-dimensional space. Active learning and/or semi-supervised learning have been explored in \cite{pelleg2005active,he2008activelearning} to iteratively interact with domain experts to label some outliers and improve the detection performance with the labeled outliers, but these methods may require extensive feedback from the experts. Different from the above contexts, our model focuses on representation learning and it incorporates a few labeled outliers by a minor change to its inputs, and is thus very flexible in practice.




\section{The Proposed Framework: \texttt{RAMODO}}

\subsection{Problem Statement}

We aim to learn a low-dimensional space out of the ultrahigh-dimensional data, such that it becomes easier and/or more efficient for random distance-based detectors to identify outliers in the new space.  Specifically, given a set of $N$ data objects $\mathcal{X}$=$\{ \mathbf{x}_{1}, \mathbf{x}_{2}, \cdots, \mathbf{x}_{N} \}$ with $\mathbf{x}_{i} \in \mathbb{R}^{D}$ (i.e., $\mathbf{x}_i$=$\{ x_{i1}, x_{i2}, \cdots, x_{iD} \}$), and a random distance-based outlier scoring function $\phi: \mathcal{X} \mapsto \mathbb{R}$ that uses distances in a random subsample to define outlierness, our goal is to learn a representation function $f: \mathcal{X} \mapsto \mathbb{R}^{M}$ ($M \ll D$) in a way that we have $\phi(f(\mathbf{x}_{i})) > \phi(f(\mathbf{x}_{j}))$ if $\mathbf{x}_{i}$ is an outlier and $\mathbf{x}_{j}$ is an inlier. 

\subsection{Ranking Model-based Representation Learning Framework}

\begin{figure}[h!]
  \centering
    \includegraphics[width=0.44\textwidth]{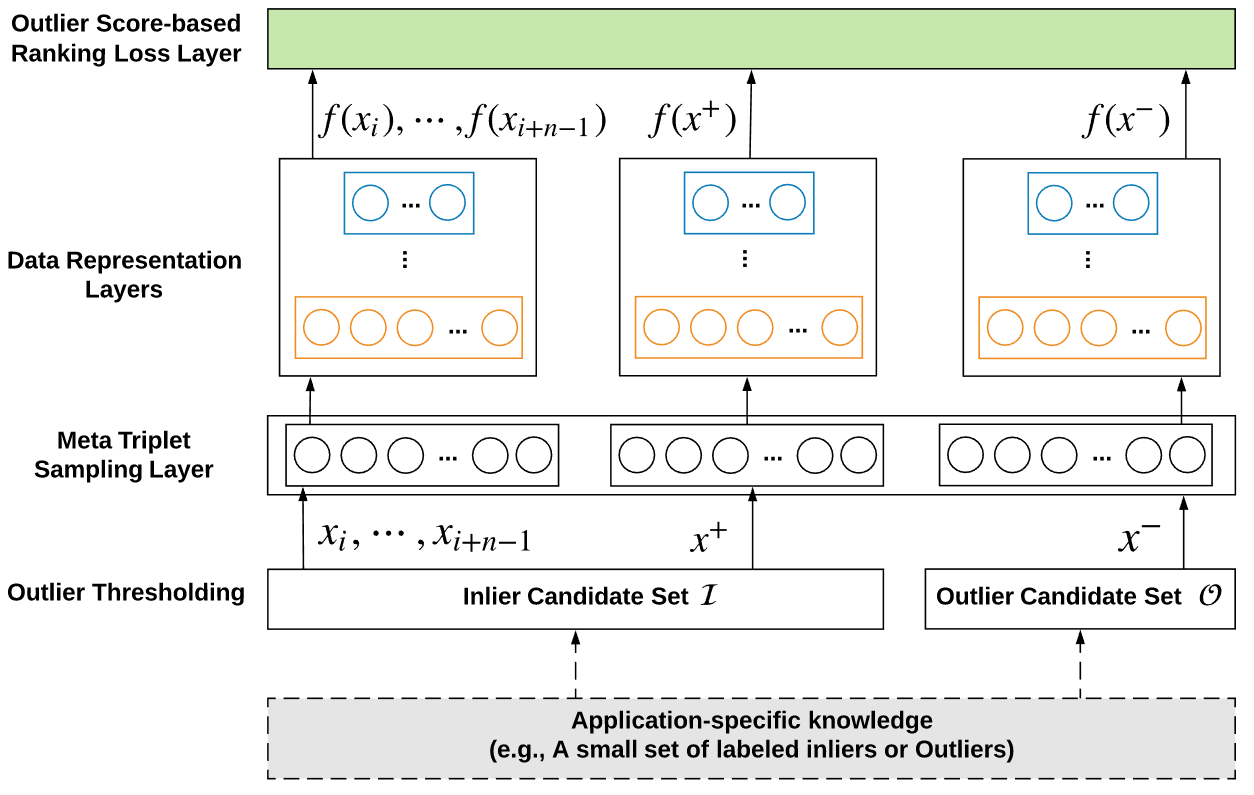}
  \caption{The Proposed RAMODO Framework. RAMODO learns a representation function $f(\cdot)$ to map $D$-dimensional input objects into a $M$-dimensional space, with $M \ll D$.}
  \label{fig:framework}
\end{figure}

We introduce the RAMODO framework by using a generalized pairwise ranking model to learn data representations that are tailored for random distance-based detectors. As shown in Figure \ref{fig:framework}, RAMODO consists of four essential components. RAMODO first performs an \textit{outlier thresholding} to partition the data into inlier and outlier candidate sets. It then generates a meta triplet sample $T=\left(<\mathbf{x}_{i}, \cdots, \mathbf{x}_{i+n-1}>, \mathbf{x}^{+}, \mathbf{x}^{-}\right)$ by randomly picking up $n$ data objects from the inlier candidate set $\mathcal{I}$ as the query set ($\mathbf{x}_{i}, \cdots, \mathbf{x}_{i+n-1}$), one object from $\mathcal{I}$ as a positive example ($\mathbf{x}^{+}$), and one object from the outlier candidate set $\mathcal{O}$ as a negative example ($\mathbf{x}^{-}$). This \textit{meta triplet sampling} serves as an input layer to a network architecture. RAMODO further learns the \textit{data representations} by a function $f$ that can be composed of one or multiple fully-connected hidden layer(s). RAMODO finally performs the optimization guided by a \textit{outlier score-based ranking loss}, $L\Big(\phi\big(f(\mathbf{x}^{+})|<f(\mathbf{x}_{i}), \cdots, f(\mathbf{x}_{i+n-1})>\big), \phi\big(f(\mathbf{x}^{-})|<f(\mathbf{x}_{i}), \cdots, f(\mathbf{x}_{i+n-1})>\big)\Big)$, in which $\phi(\cdot|\cdot)$ is a random distance-based scoring function and $L(\cdot, \cdot)$ is a loss function.

Note that RAMODO is an unsupervised framework, but it is flexible to incorporate \textit{application-specific knowledge} (e.g., some labeled inliers or outliers) into the inlier or outlier candidate set when there exists such labeled data.

\textit{Why the representations learned by RAMODO are tailored for random distance-based outlier scoring?} RAMODO optimizes the representations by encouraging $\phi\big(f(\mathbf{x}^{-})|<f(\mathbf{x}_{i}), \cdots, f(\mathbf{x}_{i+n-1})>\big)$ to be larger than $\phi\big(f(\mathbf{x}^{+})|<f(\mathbf{x}_{i}), \cdots, f(\mathbf{x}_{i+n-1})>\big)$, in which $\phi$ is equivalent to the random distance-based outlier scoring using $<f(\mathbf{x}_{i}), \cdots, f(\mathbf{x}_{i+n-1})>$ as a random subsample. The representations learned by RAMODO are therefore tailored for the random distance-based detectors. For example, $\phi$ can be instantiated to be the popular $K$NN-based detectors using the $K$-th (or average) nearest neighbor distances in the query set as outlier scores. 

\textit{How to guarantee the quality of triplet sampling?} Having a reliable outlier candidate set is the key to the quality of the triplet sampling and the representation learning. However, in an unsupervised case, we do not know whether a given $\mathbf{x}$ is an outlier or not. We show in the RAMODO's instance below that two approaches can be used to produce an outlier candidate set of good quality. The first approach is to use state-of-the-art outlier detection methods and \textit{Chebyshev}'s inequality to include the most likely outliers into the candidate set. Another approach is to incorporate a few labeled outliers into the outlier candidate set when such labeled data is available.

\textit{What are the intuitions behind RAMODO?} Since the representation function $f$ is guided by the scoring function $\phi$, $f$ only attains the information that is the most important for $\phi$ to distinguish outliers from inliers. By working on such a highly relevant space, $\phi$ is expected to obtain an accuracy that is comparable to, or better than that working on the original space even when $M \ll D$. In terms of efficiency, distance-based outlier scoring normally involves nearest neighbor searching, which can be very computationally costly in high-dimensional space since indexing methods like $k$-d tree fails to work. By working on a significantly lower-dimensional (e.g., $M \leq 30$) space, the distance computation and nearest neighbor searching in $\phi$ can be substantially accelerated when indexing methods are used.




\section{A \texttt{RAMODO} Instance: \texttt{REPEN}}

RAMODO is instantiated to the method REPEN that learns data representations for the random nearest neighbor distance-based detector, Sp \cite{sugiyama2013dbsubsample}. While Sp is a highly scalable outlier detector with significant accuracy improvement over several other popular distance-based methods and state-of-the-art high-dimensional methods \cite{sugiyama2013dbsubsample,pang2015lesinn,ting2017defying}, it still suffers from the curse of dimensionality. REPEN is customized for Sp to substantially improve its performance on ultrahigh-dimensional data.

\subsection{Outlier Thresholding Using State-of-the-art Detectors and Cantelli's Inequality}

We first use Sp to obtain an initial fairly good outlier ranking. Note that the original Sp \cite{sugiyama2013dbsubsample} may perform unstably since it only uses one single subsample to define the outlierness. Following \cite{pang2015lesinn,ting2017defying}, we use bootstrap aggregating to produce a bagging ensemble of Sp to obtain a more stable and reliable initial outlier ranking. While REPEN only uses the Sp-based ensemble for efficiency consideration, we may combine Sp with other methods to build heterogeneous ensembles \cite{rayana2016lessismore} to further improve this initial outlier ranking. 


\begin{definition}[Sp-based Outlier Scoring]
Given a data object $\mathbf{x}_{i}$, Sp defines its outlierness as follows:
\begin{equation}\label{eqn:knnscore}
    r_{i} = \frac{1}{m}\sum_{j=1}^{m} \mathit{nn}\_\mathit{dist}(\mathbf{x}_{i}|\mathcal{S}_{j}),
\end{equation}
where $\mathcal{S}_{j} \subset \mathcal{X} $ is a random data subsample, $m$ is the ensemble size, and $\mathit{nn}\_\mathit{dist}(\cdot|\cdot)$ returns the nearest neighbor distance of $\mathbf{x}_{i}$ in $\mathcal{S}_{j}$.
\end{definition}

We then use the \textit{Cantelli}'s inequality \cite{dubhashi2009concentration}, a one-sided \textit{Chebyshev}'s inequality, to define pseudo outliers. 

\begin{definition}[Cantelli's Inequality-based Outlier Thresholding]
Given an outlier score vector $\mathbf{r} \in \mathbb{R}^{N}$, in which large values indicate high outlierness, and let $\mu$ and $\delta^{2}$ be its expected value and variance, then \textit{the outlier candidate set} $\mathcal{O}$ is defined as:
\begin{equation}\label{eqn:cantelli}
    \mathcal{O} = \{\mathbf{x}_i| r_{i} \geq \mu + \alpha \delta \}, \; \forall \mathbf{x}_i \in \mathcal{X} , r_i \in \mathbf{r},
\end{equation}
where $\alpha \geq 0$ is user-defined based on a desired false positive bound.
\end{definition}

We show in Section \ref{subsec:triplet} that Eqn. (\ref{eqn:cantelli}) is equivalent to selecting an outlier candidate set with a false positive upper bound of $\frac{1}{1+\alpha^2}$. After determining $\mathcal{O}$, we obtain the \textit{inlier candidate set} by $\mathcal{I} = \mathcal{X} \setminus \mathcal{O}$. 

\subsection{Triplet Sampling Based on Outlier Scores}
An outlier score-based importance sampling is then used to generate meta triplet samples. We first sample $n$ query objects from the inlier candidate set $\mathcal{I}$ according to their outlier scores. The probability of a data object $\mathbf{x}_{i} \in \mathcal{I}$ being sampled as the query object is inversely proportional to its outlier score and is defined as follows:
\begin{equation}\label{eqn:sampling_query}
    p(\mathbf{x}_i) = \frac{\mathbb{Z} - r_{i}}{\sum_{t=1}^{|\mathcal{I}|}[\mathbb{Z} - r_{t}]},
\end{equation}
\noindent where $\mathbb{Z} = \sum_{t=1}^{|\mathcal{I}|} r_{t}$. The importance sampling offers high probability of choosing a set of representative inliers as query objects. 

We then sample a positive example $\mathbf{x}^{+}$ from $\mathcal{I}$ using uniform sampling. Instead of importance sampling, we use uniform sampling to diversify the positive examples in different triplets.

We further sample a negative example $\mathbf{x}^{-}$ from the outlier candidate set $\mathcal{O}$. Importance sampling is used here to obtain high probability of choosing the most likely outliers as negative examples. Given a data object $x_{j} \in \mathcal{O}$, its probability of being chosen as a negative example is defined as:
\begin{equation}\label{eqn:sampling_negative}
    p(\mathbf{x}_j) = \frac{r_{j}}{\sum_{t=1}^{|\mathcal{O}|} r_{t}}.
\end{equation}

\subsection{A Shallow Data Representation}
One single hidden layer is defined below to learn a shallow data representation. The shallow representation is used for two main reasons. (i) In many ultrahigh-dimensional data, we often have $N \ll D$. As a result, we may not have sufficient data to train a deep representation. (ii) Deep representation learning requires extensive computation for data sets with millions of features.

\begin{definition}[Single-layer Fully-connected Representations]
Given an input $\mathbf{x}$, it is mapped to a new space of $M$ dimensions by:
\begin{equation}\label{eqn:representation}
    f_{\Theta}(\mathbf{x}) = \{\psi(\mathbf{w}_{1}^{\intercal}\mathbf{x}), \psi(\mathbf{w}_{2}^{\intercal}\mathbf{x}), \cdots,\psi(\mathbf{w}_{M}^{\intercal}\mathbf{x} ) \},
\end{equation}
where $\psi(\cdot)$ is an activation function, $\mathbf{w}_{i} \in \mathbb{R}^{D}$ is a weight vector, and $\Theta=\{\mathbf{w}_{1},\mathbf{w}_{2},\cdots,\mathbf{w}_{M}\}$ is the parameter set to be learned.
\end{definition}

The ReLu function $\psi(z) = \mathit{max}(0, z)$ is used because of its efficient computation and gradient propagation. 

\subsection{Ranking Loss Using Random Nearest Neighbor Distance-based Outlier Scores}

A random nearest neighbor distance-based function is used in $\phi(\cdot|\cdot)$ for the loss function $L(\cdot, \cdot)$ to learn customized data representations for Sp. Let $\mathcal{Q} = <f_{\Theta}(\mathbf{x}_{i}), \cdots, f_{\Theta}(\mathbf{x}_{i+n-1})>$ be the query set, then given an object $\mathbf{x}$, REPEN defines the outlierness of $f_{\Theta}(\mathbf{x})$ using its nearest neighbor distance in $\mathcal{Q}$:
\begin{equation}
\phi \big(f_{\Theta}(\mathbf{x})|\mathcal{Q} \big) = \mathit{nn}\_\mathit{dist}\big(f_{\Theta}(\mathbf{x})|\mathcal{Q} \big).
\end{equation}
Hence, given a triplet $T=\big(\mathcal{Q}, f_{\Theta}(\mathbf{x}^{+}), f_{\Theta}(\mathbf{x}^{-}) \big)$, our goal is to learn a representation function $f(\cdot)$ that results in 
\begin{equation}\label{eqn:goal}
  \mathit{nn}\_\mathit{dist}\big(f_{\Theta}(\mathbf{x}^{+})|\mathcal{Q} \big) < \mathit{nn}\_\mathit{dist}\big(f_{\Theta}(\mathbf{x}^{-})|\mathcal{Q}\big), 
\end{equation}
i.e., the pseudo outlier $\mathbf{x}^{-}$ has a larger nearest neighbor distance in $\mathcal{Q}$ than the pseudo inlier $\mathbf{x}^{+}$. We then define the following hinge loss function for the triplet $T$ to achieve this goal:
\begin{multline}
    J(\Theta; T) = L\Big (\phi \big(f_{\Theta}(\mathbf{x}^{+})|\mathcal{Q} \big), \phi \big( f_{\Theta}(\mathbf{x}^{-})|\mathcal{Q} \big) \Big) = \\ \max{\{0, c + \mathit{nn}\_\mathit{dist}\big(f_{\Theta}(\mathbf{x}^{+})|\mathcal{Q} \big) - \mathit{nn}\_\mathit{dist}\big(f_{\Theta}(\mathbf{x}^{-})|\mathcal{Q}\big)\}},
\end{multline}
\noindent where $c$ is a margin parameter that controls the difference between the two distances $\mathit{nn}\_\mathit{dist}\big(f_{\Theta}(\mathbf{x}^{+})|\mathcal{Q} \big)$ and $\mathit{nn}\_\mathit{dist}\big(f_{\Theta}(\mathbf{x}^{-})|\mathcal{Q}\big)$. The hinge loss is a convex function, which penalizes the violation of the ranking order in Eqn. (\ref{eqn:goal}) and encourages a separation between positive examples (inliers) and negative examples (outliers). 

Given a set of triplets $\mathcal{T}$, our objective function becomes
\begin{equation}
    \argmin_{\Theta} \frac{1}{|\mathcal{T}|} \sum_{i=1}^{|\mathcal{T}|} J(\Theta; T_{i}).
\end{equation}

Since the positive and negative examples come from the bottom-ranked and top-ranked objects of the ranking $\mathbf{r}$ respectively, given an infinite number of $T$, REPEN optimizes the precision at top $|\mathcal{O}|$ of the detector Sp by minimizing the loss function $J$. Additionally, REPEN does not learn the representation dimension $M$ automatically, since there does not exist reliable supervision information to effectively guide the learning.

\subsection{The Algorithm and Its Time Complexity}\label{subsec:algorithm}

Algorithm \ref{alg:repend} presents the procedure of REPEN. Step 1 uses an ensemble of Sp to produce an outlier ranking. Following \cite{sugiyama2013dbsubsample,pang2015lesinn}, a small subsampling size of 8 and an ensemble size of 50 are used to yield an initial reliable ranking $\mathbf{r}$. Step 2 yields the inlier and outlier candidate sets using the \textit{Cantelli}'s inequality with $\alpha=1.732$, which is equivalent to an outlier thresholding with a 25\% false-positive upper bound. After a random weight initialization in Step 3, REPEN iteratively learns the parameters $\Theta$ using mini-batch gradient descents in Steps 4-13. 

Specifically, Steps 6-9 generate one mini-batch of training triplets of size $b$. We found that REPEN using $n=1$ performs stably, and more importantly, runs substantially faster than that using a larger $n$. According to the analysis in Section \ref{subsec:upperbound}, using small $n$ can also result in a small error bound of the representation learning. $n=1$ is thus used. The parameter settings of $n\_epochs$ , $n\_batches$ and $b$ are provided in Section \ref{subsubsec:realworlddata_settings}. Additionally, the sampling in Steps 6-9 is with replacement. Step 10 then computes the average loss given one batch of triplet samples, in which square Euclidean distance is used in $\mathit{nn}\_\mathit{dist}(\cdot|\cdot)$ and $c=1000$ is used in order to encourage large margins between inliers and outliers in the representation space. After that, Step 11 computes the partial derivatives w.r.t. the weight parameters and performs a gradient descent step using the ADADELTA optimization \cite{zeiler2012adadelta}. Lastly, Step 14 uses the function $f$ to map all objects into the representation space. 


\renewcommand{\algorithmicrequire}{\textbf{Input:}}
\renewcommand{\algorithmicensure}{\textbf{Output:}}
\begin{algorithm}
\small 
\caption{\textit{REPEN}}
\begin{algorithmic}[1]
\label{alg:repend}
\REQUIRE $\mathcal{X} \in \mathbb{R}^{D}$ - ultrahigh-dimensional data objects
\ENSURE $\mathcal{X}^{\prime} \in \mathbb{R}^{M}$ - low-dimensional data representations
\STATE $\mathbf{r} \leftarrow \frac{1}{m}\sum_{i=1}^{m}\mathit{nn}\_\mathit{dist}(\mathcal{X}|\mathcal{S}_{i})$
\STATE $\mathcal{O} \leftarrow \{\mathbf{x}_i| r_{i} \geq \mu + a \delta \}, \; \forall \mathbf{x}_i \in \mathcal{X} $, and $\mathcal{I} \leftarrow \mathcal{X} \setminus \mathcal{O}$
\STATE Randomly initialize $\Theta$
\FOR{ $i = 1$ to $n\_epochs$}
    \FOR{ $j = 1$ to $n\_batches$}
        \STATE $\mathcal{B}_{q} \leftarrow$ Sample $\mathit{b}$ sets of $n$-sized query set from $\mathcal{I}$ using Eqn. (\ref{eqn:sampling_query})
        \STATE $\mathcal{B}_{p} \leftarrow$ Sample $\mathit{b}$ objects from $\mathcal{I}$ using uniform sampling
        \STATE $\mathcal{B}_{n} \leftarrow$ Sample $\mathit{b}$ objects from $\mathcal{O}$ using Eqn. (\ref{eqn:sampling_negative})
        \STATE Generate the triplet set $\mathcal{T}=\{T_{1},\cdots,T_{b}\}$, with each triplet $T_{s}=\left (<\mathbf{x}_{t}, \cdots, \mathbf{x}_{t+n-1}>, \mathbf{x}^{+}, \mathbf{x}^{-} \right)$ from the $s$-th element of $\mathcal{B}_{q}$, $\mathcal{B}_{p}$, and $\mathcal{B}_{e}$ respectively
        \STATE $J(\Theta) \leftarrow \frac{1}{b}\sum_{s=1}^{b}J(\Theta; T_{s})$    
        \STATE Perform a gradient descent step w.r.t. the parameters in $\Theta$
    \ENDFOR
\ENDFOR
\STATE $\mathcal{X}^{\prime} \leftarrow f_{\Theta}(\mathcal{X})$
\RETURN $\mathcal{X}^{\prime}$
\end{algorithmic}
\end{algorithm}

Step 1 requires $O(mND|\mathcal{S}|)$ to obtain the nearest neighbor distance-based outlier scores for all objects. Step 2 requires $O(N)$ to scan over the outlier ranking list to produce the inlier and outlier candidate sets. Since $n\_epochs$, $n\_batches$ and $b$ are small constant, the time complexity of the optimization in Steps 4-13 is mainly determined by the computation of the loss function in Step 9 and the gradient descents in Step 10. Since REPEN has only one fully connected hidden layer, both the loss function and gradient descents have a time complexity linear w.r.t. $D$ and $M$. After the optimization, REPEN takes $O(NDM)$ to map the original data into the representation space. The overall time complexity of REPEN is expected to be linear w.r.t. data size and dimensionality size.

\subsection{Leveraging A Few Labeled Outliers to Improve Triplet Sampling}\label{subsec:knowledge}

Given a small set of $l$ labeled outliers, a minor change to the sampling of negative examples in REPEN (i.e., Step 8 in Algorithm \ref{alg:repend}) can well leverage them to improve the quality of triplets. Specifically, we first sample $\frac{b}{2}$ objects from the outlier candidate set $\mathcal{O}$ using the importance sampling described in Eqn. (\ref{eqn:sampling_negative}); and then the other half of the negative examples are sampled from the set of $l$ labeled outliers using uniform sampling with replacement. The negative examples from $\mathcal{O}$ may cover types of outliers that do not present in the labeled outliers, while sampling from the labeled outlier pool guarantees the presence of outliers in the negative examples. Hence, these two sources of negative examples can well complement each other for more effective subsequent representation learning.

Note that the above method assumes the availability of only a few labeled outliers since it is typically very difficult or costly to collect labeled outliers. We may sample all negative examples from the labeled outliers in the cases where the number of labeled outliers is sufficiently large to cover different types of outliers.

\section{Theoretical Foundation of \texttt{REPEN}}

\subsection{Upper Error Bound for the Representation Learning}\label{subsec:upperbound}
We show below that REPEN can obtain an upper error bound for its representation learning based on the theorems in \cite{sugiyama2013dbsubsample}. An equivalent form of $K$-th nearest neighbor distance-based outliers is the $\mathit{DB}(\beta, \delta)\text{-outliers}$ \cite{knox1998dboutlier} that defines an object $\mathbf{x}$ as an outlier if $|\{\mathbf{x}^{\prime} \in \mathcal{X}| \mathit{dist}(\mathbf{x}, \mathbf{x}^{\prime}) > \delta \} | \geq \beta N$, where $\mathit{dist}$ is a distance measure. We denote the $\mathit{DB}(\beta, \delta)\text{-outliers}$ by $\mathcal{O}(\beta; \delta)$, and let $\mathcal{I}(\beta; \delta) = \mathcal{X} \setminus \mathcal{O}(\beta; \delta)$ be the rest of instances. $\mathcal{I}_{s}(\beta; \delta)$ is a subset of $\mathcal{I}(\beta; \delta)$ such that $\min_{\mathbf{x}^{\prime} \in \mathcal{I}_{s}(\beta; \delta)} \mathit{dist}(\mathbf{x},\mathbf{x}^{\prime}) > \delta$ for every outlier $\mathbf{x} \in \mathcal{O}(\beta; \delta)$. Let a $\delta$-radius partition $\mathcal{P}_{\delta}$ of $\mathcal{I}_{s}(\beta; \delta)$ be a set of $h$ non-empty disjoint clusters, i.e., $\mathcal{I}_{s}(\beta; \delta)=\cup_{i=1}^{h}\mathcal{C}_{i}$, such that each cluster satisfies $\max_{ \mathbf{x}, \mathbf{x}^{\prime} \in \mathcal{C}  }\ \mathit{dist}(\mathbf{x}, \mathbf{x}^{\prime}) < \delta$. It shows in \cite{sugiyama2013dbsubsample} that
\begin{multline}
    p\left(\mathit{nn}\_\mathit{dist}(\mathbf{x}|\mathcal{S}) \leq \mathit{nn}\_\mathit{dist}(\mathbf{x}^{\prime}|\mathcal{S}) \right) < \\ 1 - \gamma^{n}\max_{\mathcal{P}_{\delta}}\eta_{\delta}(n), \; \forall \mathbf{x} \in \mathcal{O}(\beta; \delta), \forall \mathbf{x}^{\prime} \in \mathcal{I}(\beta; \delta),
\end{multline}
\noindent $\gamma = \frac{|\mathcal{I}_{s}(\beta; \delta)|}{N}$ and $\eta_{\delta}(n)=\sum_{\forall i;n_{i} \gneq 0}g(n_{1},\cdots,n_{h};n,p_1,\cdots,p_{h})$, in which $n=|\mathcal{S}|$ and $g$ is a probability mass function of the multinomial distribution with $\sum_{i=1}^{h}n_{i}=n$ and $\sum_{i=1}^{h}p_{i}=1$. By replacing $\mathbf{x}$ with $f_{\Theta}(\mathbf{x})$ and $\mathcal{S}$ with $\mathcal{Q}$, we obtain the following error bound:
\begin{equation}\label{eqn:errorbound}
    p\Big(\mathit{nn}\_\mathit{dist}\big(f_{\Theta}(\mathbf{x}^{-})|\mathcal{Q}\big) \leq \mathit{nn}\_\mathit{dist}\big(f_{\Theta}(\mathbf{x}^{+})|\mathcal{Q}\big) \Big) <  1 - \gamma^{n}\max_{\mathcal{P}_{\delta}}\eta_{\delta}(n).
\end{equation}

Different from \cite{sugiyama2013dbsubsample} that uses $\mathit{nn}\_\mathit{dist}(\cdot|\cdot)$ to identify outliers directly, REPEN uses the difference between $\mathit{nn}\_\mathit{dist}\big(f_{\Theta}(\mathbf{x}^{+})|\mathcal{Q} \big)$ and $\mathit{nn}\_\mathit{dist}\big(f_{\Theta}(\mathbf{x}^{-})|\mathcal{Q}\big)$ to perform the gradient descent steps in its representation learning. Hence, Eqn. (\ref{eqn:errorbound}) can be seen as an upper error bound for the representation learning. 

We often have a sufficiently large $\gamma$ in Eqn. (\ref{eqn:errorbound}). This is because $\mathcal{I}_{s}(\beta; \delta)$ represents a set of inliers that demonstrate very different characteristics from outliers, and the proportion of such inliers can be very large in practice. Hence, the error bound is often a small value when using a small query set $\mathcal{Q}$ (i.e., a small $n$). In such cases, the expressiveness of REPEN's representations is well guaranteed when the pseudo outliers $\mathbf{x}^{-}$ and inliers $\mathbf{x}^{+}$ are genuine.


\subsection{Reliable Triplet Sampling}\label{subsec:triplet}
Since it has a high probability of hitting genuine inliers given their large proportion, having genuine outliers in the outlier candidate set is the key to the representation learning. Thus, the above error bound has a good guarantee when a few labeled outliers are available. In the case of only having the score vector $\mathbf{r}$, our problem is to determine an outlierness threshold for $\mathbf{r}$, such that we obtain an outlier candidate set with small false positives. We show below that the \textit{Cantelli}'s inequality can help achieve this goal. Given $\mu$ and $\delta^{2}$ be the mean and variance of $\mathbf{r}$, we have $p(r_{i} \geq \mu + \epsilon ) \leq \frac{\delta^2}{\delta^2 +\epsilon^2}$ with $\epsilon \geq 0$ per the \textit{Cantelli}'s inequality. By replacing $\epsilon = \alpha \delta$, we obtain 
\begin{equation}
    p(r_i \geq  \mu + \alpha \delta ) \leq \frac{1}{1+\alpha^2},
\end{equation}

\noindent which states that most $\mathbf{r}$ values distribute very closely to the expected value, with the probability of up to $\frac{1}{1+\alpha^2}$ that a few exceptions occur. In other words, when $\mu + \alpha \delta$ is used as the threshold of labeling outliers, we have a false positive upper bound of $\frac{1}{1+\alpha^2}$. Note that the \textit{Cantelli}'s inequality holds for any distribution that has statistical mean and variance, and thus makes no assumption on specific probability distributions. This property enables us to obtain a good outlier candidate set even when $\mathbf{r}$ follows different distributions in different data sets.

\section{Experiments}

\subsection{Data Sets}
As shown in Table \ref{tab:aucandtime}, eight real-world ultrahigh-dimensional data sets are used, which cover diverse domains, e.g., malicious URL detection, cancer detection, molecular bioactivity detection, and Internet advertisement detection. OvarianTumour (\textit{OT}) is taken from the RSCTC Discovery Challenge for medical diagnosis and treatment\footnote{http://tunedit.org/challenge/RSCTC-2010-A}. \textit{Webspam} is taken from the Pascal Large Scale Learning Challenge\footnote{ftp://largescale.ml.tu-berlin.de/largescale/}. \textit{URL} consists of 120-day collection of malicious and benign URLs \cite{ma2009url}. Since \textit{URL} contains evolving URLs data, the first-week subset of this collection is used. \textit{R8}\footnote{http://csmining.org/tl\_files/Project\_Datasets/r8\_r52/r8-train-all-terms.txt} and  \textit{News20}\footnote{https://www.csie.ntu.edu.tw/$\sim$cjlin/libsvmtools/datasets/binary/news20.binary.bz2} are text classification data sets. \textit{AD} is a data set for detecting Internet advertisements, LungCancer (\textit{LC}) is for cancer detection, and \textit{p53} is about abnormal protein activity detection. These three data sets are available at UCI Machine Learning Repository\footnote{https://archive.ics.uci.edu/ml/index.php}. The data sets \textit{OT}, \textit{URL}, \textit{AD}, \textit{LC}, \textit{p53} and \textit{Webspam} contain real outliers, but \textit{URL} and \textit{Webspam} were originally prepared for balanced classification tasks. Following the literature \cite{zimek2013subsampling,campos2016evaluation,keller2012hics}, \textit{URL} and \textit{Webspam} are converted to outlier detection data sets with 2\% outliers by downsampling the small class or the positive class. This downsampling approach is also applied to \textit{News20}. For \textit{R8} that is very imbalanced, we follow \cite{lazarevic2005fb,keller2012hics,rayana2016lessismore,ting2017defying} to treat the rare classes as outliers and the largest class as inliers. After the above data preparation, all data sets contain (semantically) real outliers.

\subsection{Performance Evaluation Methods}
We evaluate the effectiveness of the representations by the performance of the subsequent outlier detector Sp. Sp is used with the recommended settings as in \cite{pang2015lesinn}. Given a set of data objects in the original space or the representation space, Sp returns a full ranking list of data objects based on their outlier scores. After that, many performance measures, such as the area under Receiver Operating Characteristic curve (AUC) and average precision, are available for evaluating the quality of the ranking. A detailed analysis of the advantages and drawbacks of these measures for unsupervised outlier detection can be found in \cite{campos2016evaluation}. Following the literature \cite{sugiyama2013dbsubsample,ting2017defying,keller2012hics,zimek2013subsampling,campos2016evaluation}, the popular measure AUC is used. AUC inherently considers the class-imbalance nature of outlier detection, making it comparable across data sets with different outlier proportions \cite{campos2016evaluation}.  An AUC value close to 0.5 indicates a random ranking of the objects. Higher AUC indicates better detection performance.

The paired \textit{Wilcoxon} signed rank test is used to examine the significance of the performance of REPEN against its competitors. For algorithms or data sets that involve sampling, their AUCs are the averaged results over 10 independent runs.

The runtime presented in our experiments is calculated at a node in a 2.8GHz Titan cluster with 256GB memory.

\subsection{Effectiveness in Real-world Data with Thousands to Millions of Features}\label{subsec:realworlddata}

\subsubsection{Experiment Settings}\label{subsubsec:realworlddata_settings}

We compare the AUC performance and detection runtime of Sp in the low-dimensional representation space learned by REPEN and in the original high-dimensional space to evaluate the effectiveness of REPEN. Sp is applied with the same settings, subsampling size set to 8 and ensemble size set to 50, in both the representation space and original space. For the nearest neighbor searching, $k$-d tree indexing is used. 

REPEN with the default representation dimension $M=20$ is used. In the optimization, $n\_epochs = 30$, $b=256$ and 5,000 samples per epoch (equivalent to $\frac{5,000}{b}$ batches per epoch) are used by default. Note that Sp is very cost-effective, which only requires the presence of a few representative inliers in a subsample to effectively distinguish outliers from inliers \cite{sugiyama2013dbsubsample,pang2015lesinn,ting2017defying}. Therefore, although the number of possible triplets is huge for large data sets, our experiments showed that these settings enable the optimization to achieve the convergence and impressive detection accuracy for all the data sets except \textit{p53}, in which outliers are deeply mixed with inliers and thus we increase the number of samples per epoch to its data size to sufficiently train the representation model. Both Sp and REPEN are implemented in Python \footnote{The code of REPEN is available at https://sites.google.com/site/gspangsite/sourcecode.}. 

\subsubsection{Findings - REPEN Enables the Detector Sp to Obtain Significantly Better AUC Performance and Two Orders of Magnitude Speedup}

Table \ref{tab:aucandtime} shows the AUC performance and \textit{detection runtime}\footnote{The detection runtime is the execution time of online detection, which does not include the data loading and offline training time.} of Sp in the original data space and the 20-D space output by REPEN. Although REPEN retains only 20 dimensions, which are at most 1.3\% of the dimensionality size in the original data, the REPEN-enabled Sp is either substantially better than, or roughly the same as, that working on the original space. Our significance test based on the AUC performance on the eight data sets obtains a p-value of 0.0156, indicating that the REPEN-enabled Sp significantly outperforms the original Sp at the 95\% confidence level. 

\begin{table}[htbp]
  \centering
  \caption{AUC and Detection Runtime of the Original Sp (ORG) and the REPEN-enabled Sp (REPEN). $D$ is the dimensionality size and $N $ is the data size. \textit{IMP} and \textit{SU} indicate the AUC improvement and speedup of REPEN over ORG, respectively.}
  \scalebox{0.80}{
    \begin{tabular}{|p{1.1cm}p{1.1cm}p{0.8cm}|p{0.7cm}p{0.77cm}|c|p{0.8cm}p{0.75cm}|c|}
    \hline
       &    &    & \multicolumn{3}{|c}{\textbf{AUC}} & \multicolumn{3}{|c|}{\textbf{Runtime (s)}} \\
    \hline
    Data & $D$  & $N $ & ORG & REPEN & \textit{IMP} & ORG & REPEN & \textit{SU} \\\hline
    AD & 1,555 & 3,279 & 0.7117 & \textbf{0.8533} & 19.90\% & 0.16 & 0.10 & 2 \\
    LC & 3,312 & 145 & 0.9149 & \textbf{0.9283} & 1.47\% & 0.59 & 0.02 & 38 \\
    p53 & 5,408 & 16,772 & 0.6686 & \textbf{0.6817} & 1.95\% & 230.39 & 0.91 & 253 \\
    R8 & 9,467 & 3,974 & 0.8602 & \textbf{0.9080} & 5.56\% & 0.21 & 0.12 & 2 \\
    OT & 54,621 & 265 & 0.6976 & \textbf{0.7627} & 9.34\% & 23.50 & 0.04 & 529 \\
    News20 & 1,355,191 & 10,201 & 0.5361 & \textbf{0.5822} & 8.60\% & 5.30 & 0.29 & 18 \\
    URL & 3,231,961 & 89,063 & 0.7556 & \textbf{0.7733} & 2.35\% & 16.40 & 2.48 & 7 \\
    Webspam & 16,609,143 & 215,001 & \textbf{0.8781} & 0.8713 & -0.78\% & 1879.68 & 6.08 & 309 \\
    \hline
    \end{tabular}%
    }
  \label{tab:aucandtime}%
\end{table}%

Particularly, REPEN enables Sp to obtain about 1.5\% to 20\% AUC improvement on most data sets. This is mainly due to the fact that there often exists only a small percentage of features that are relevant to outlier detection, since outliers are the minority objects; the large proportion of irrelevant features in the original space renders Sp less effectively. In contrast, the customized objective function enables REPEN to effectively preserve the most important information for Sp. Therefore, Sp performs outlier detection in a highly relevant space when working with our representation space, resulting in the above substantial improvement.

In terms of runtime, by working on our 20-D representation space, Sp runs up to two-order of magnitude faster than on the original space. This is because the original data has very high dimensionality, data indexing methods fail to work. In contrast, $k$-d tree indexing excels at the 20-D representations, which enables Sp to achieve a remarkable speedup. Such speedup is striking for the two dense data sets, \textit{p53} and \textit{OT}, and the large sparse data set, \textit{Webspam}. Note that we have used the efficient implementation of distance measures that is specifically designed for high-dimensional sparse data in Sp, so Sp runs very fast in the small sparse data sets, \textit{News20} and \textit{URL}.

\subsection{Comparing to State-of-the-art Representation Learning Competitors}

\subsubsection{Experiment Settings}
This section compares REPEN with autoencoder (AE) \cite{hinton2006ae}, Hessian locally linear embedding (HLLE) \cite{donoho2003hessian}, sparse random projection (SRP) \cite{li2006srp}, and coherent pursuit (CoP) \cite{rahmani2017cop}. These methods are chosen because they are state-of-the-art methods of four popular unsupervised representation learning approaches, i.e., neural network-based approaches, manifold learning, random projections, and robust PCA approaches. All methods learn a mapping from the original space to a space of the same dimension ($M=20$) to examine their effectiveness of learning low-dimensional representations. AE uses the same training settings as REPEN. Since HLLE requires its neighborhood size $K>\frac{M(M+3)}{2}$ in its optimization, $K=250$ is used. SRP and CoP do not have additional parameters. AE, HLLE and SRP are implemented using the built-in functions in the Scikit-learn Python library. CoP is taken from their authors in MATLAB.

\subsubsection{Findings - REPEN Performs Substantially Better and More Stably Than the Competitors}

Table \ref{tab:competitor} demonstrates the AUC performance of Sp using the representations output by REPEN, AE, HLLE, SRP, and CoP. REPEN performs significantly better than AE and SRP at the 95\% confidence level, and is comparably better than HLLE and Cop. In general, REPEN enables Sp to achieve substantial improvement over its competitors on the data sets \textit{AD}, \textit{R8}, \textit{OT}, \textit{URL}, and \textit{Webspam}, in which Sp has fairly good performance on the original space. For example, the REPEN-based Sp obtains about 14\% to 32\% AUC improvement over all its competitors on \textit{AD}, 5\% to 42\% improvement over all its competitors on \textit{OT}, and 27\% to 33\% improvement over AE and SRP on \textit{URL}. This is because when Sp obtains good performance on the original data space, the triplet sampling in REPEN yields higher-quality training triplet samples, which lead its optimization to better representations for Sp. The four competitors learn representations independently of the detector, and thus they are unable to achieve this merit. Note that the representation learning in the methods AE and SRP is simpler than the other methods, and they often require a sufficiently large number of representation features (e.g., hundreds to thousands of features) to perform fairly well. Therefore, they perform poorly in most data sets when they only retain 20 representation features.

\begin{table}[htbp]
  \centering
  \caption{AUC Performance of Sp using REPEN, AE, HLLE, SRP, and CoP. `$\bullet$' indicates out-of-memory errors while `$\circ$' denotes algorithmic constraint violation errors.}
\vspace{-2mm}
  \scalebox{0.85}{
    \begin{tabular}{|lccccc|}
    \hline
    \textbf{Data} & \textbf{REPEN} & \textbf{AE} & \textbf{HLLE} & \textbf{SRP} & \textbf{CoP} \\
    \hline
    AD & \textbf{0.8533} & 0.6848 & 0.6462 & 0.7249 & 0.7483 \\
    LC & \textbf{0.9283} & 0.9266 & $\circ$ & 0.9041 & 0.9161 \\
    p53 & \textbf{0.6817} & 0.5624 & 0.3870 & 0.6874 & 0.6805 \\
    R8 & 0.9080 & 0.7698 & 0.8764 & 0.8494 & \textbf{0.9326} \\
    OT & \textbf{0.7627} & 0.5355 & 0.7273 & 0.6708 & 0.7247 \\
    News20 & 0.5822 & 0.5154 & \textbf{0.6206} & 0.4435 & $\bullet$ \\
    URL & \textbf{0.7733} & 0.6040 & $\bullet$ & 0.5774 & $\bullet$ \\
    Webspam & 0.8713 & \textbf{0.8766} & $\bullet$ & 0.6390 & $\bullet$ \\\hline
    \multicolumn{2}{|r}{p-value}    & 0.0234 & 0.3125 & 0.0156 & 0.3125 \\
    \hline
    \end{tabular}%
    }
  \label{tab:competitor}%
\end{table}%

Also, REPEN performs much more stably than the other methods across the data sets. REPEN obtains the best performance in five data sets, with the other three ranked in second. Although AE, HLLE and CoP respectively obtain one best performance, they perform poorly in several other data sets, e.g., the performance of AE and HLLE on \textit{AD} and \textit{p53}. One main reason here is that AE, HLLE and CoP are not specifically designed for the outlier detectors, and as a result, the best representations they obtain may be suboptimal for Sp. This reinforces the importance of incorporating the subsequent detection methods into the representation learning.

CoP outperforms REPEN on \textit{R8}. This may be because CoP is able to preserve information for detecting some structural outliers in \textit{R8}, while REPEN fails to do that. HLLE outperforms REPEN on \textit{News20}. This may be because a large neighborhood size is required to detect more outliers than REPEN does. We plan to instantiate other RAMODO's instances to handle these two issues in future.

\subsection{The Capability of Leveraging Labeled Outliers as Prior Knowledge}\label{subsec:priorknowledge}

\subsubsection{Experiment Settings}

This section examines the capability of REPEN in incorporating a small set of labeled outliers into its representation learning. Since we downsampled one class in \textit{News20}, \textit{URL}, and \textit{Webspam} to create outliers, the rest of data objects in the downsampled class can be used as the pool of labeled outliers. We randomly pick up $l$ objects from the pool and add them into our training using the method described in Section \ref{subsec:knowledge}. $l \in \{1, 5, 10, 20, 40, 80\}$ is used. The AUC performance of using a set of $l$ labeled outliers is averaged over 10 independent runs.

\subsubsection{Findings - REPEN Achieves Up to 32\% AUC Improvement by Leveraging 1-80 Labeled Outliers}

\begin{figure}[h!]
  \centering
    \includegraphics[width=0.42\textwidth]{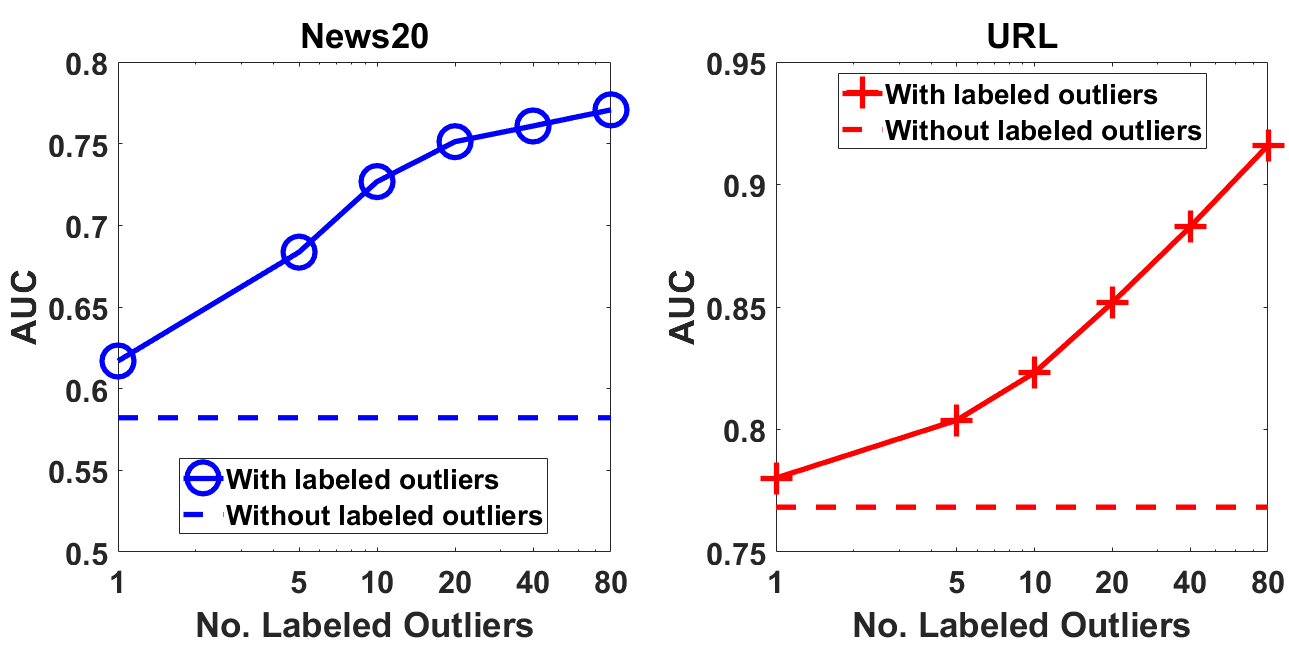}
  \caption{AUC Results of REPEN-based Sp Using Labeled Outliers.}
  \label{fig:dk}
\end{figure}

Figure \ref{fig:dk} illustrates the AUC performance of REPEN on \textit{News20} and \textit{URL} using different numbers of labeled outliers. Similar results are observed on \textit{Webspam}.  The results of REPEN on \textit{News20} and \textit{URL} in Table \ref{tab:competitor} are used as the baselines, i.e., the performance of REPEN without using the labeled outliers. The AUC performance increases quickly with the increasing of the number of known outliers. For the complex data set \textit{News20}, in which REPEN originally obtains an AUC of only 0.5822, REPEN can harness the limited number of available outliers to improve the AUC up to 0.7707. For the easier data set \textit{URL}, REPEN improves the AUC performance from 0.7733 to 0.9160 when the labeled outliers are provided. This is mainly because the labeled outlier-based prior knowledge largely improves our training data quality and helps identify more application-relevant outliers.

We further investigate the changes of outlier statistics of the two data sets w.r.t. the number of labeled outliers. The results are reported in Table \ref{tab:news20andurl}. The number of labeled data objects is less than 1\% of the unlabeled data according to the $\frac{l}{N}$ factor. Also, the proportion of (labeled and unlabeled) outliers (i.e., $\frac{l+N_{o}}{N}$) remains to be very small for both data sets. These two data factors indicate that the intrinsic complexity of the problem does not decrease due to the available of a few labeled outliers. The third factor $\frac{l}{N_{o}}$ represents the ratio of the number of the labeled outliers to the number of unlabeled outliers in the data sets. For \textit{URL}, REPEN can use the labeled outliers that account for only up to $4.50\%$ of unlabeled outliers to achieve 18\% AUC improvement. The $\frac{l}{N_{o}}$ in \textit{News20} is much larger than that in \textit{URL}. In such cases, REPEN leverages the extra labeled outliers to achieve more substantial improvement, resulting in more than 32\% AUC improvement. These results show the strong capability of REPEN in making very effective use of the small number of labeled outliers in different cases.

\begin{table}[htbp]
  \centering
  \caption{Outlier Statistics of News20 and URL w.r.t. the Number of Labeled Outliers. $l$ is the number of labeled outliers, $N_{o}$ is the number of unlabeled outliers in the data, and $N$ is the data size.}
\vspace{-2mm}
  \scalebox{0.85}{
    \begin{tabular}{|c|ccccccc|}
    \hline
     Factor  & Data   & 1  & 5  & 10 & 20 & 40 & 80 \\
    \hline
    \multirow{2}[0]{*}{$\frac{l}{N}$} & News20  & 0.01\% & 0.05\% & 0.10\% & 0.20\% & 0.39\% & 0.78\%  \\
       & URL & 0.00\% & 0.01\% & 0.01\% & 0.02\% & 0.04\% & 0.09\%  \\\hline
    \multirow{2}[0]{*}{$\frac{l+N_{o}}{N}$} & News20 & 2.01\% & 2.05\% & 2.10\% & 2.20\% & 2.39\% & 2.78\%  \\
       & URL & 2.00\% & 2.00\% & 2.01\% & 2.02\% & 2.04\% & 2.09\% \\\hline
    \multirow{2}[0]{*}{$\frac{l}{N_{o}}$} & News20 & 0.49\% & 2.45\% & 4.90\% & 9.80\% & 19.61\% & 39.22\%  \\
       & URL & 0.06\% & 0.28\% & 0.56\% & 1.13\% & 2.25\% & 4.50\% \\
    \hline
    \end{tabular}%
    }
  \label{tab:news20andurl}%
\end{table}%


\subsection{Sensitivity Test w.r.t. the Representation Dimension}

\subsubsection{Experiment Settings}

We examine the sensitivity of the performance of REPEN w.r.t. the number of representation features, M, on all the eight data sets. A wide range of M values, $\{1, 10, 20, 30, 40, 50, 60, 70, 80, 90, 100\}$, is used.

\subsubsection{Findings - REPEN Performs Stably w.r.t. Different Numbers of Representation Features}

\begin{figure}[h!]
  \centering
    \includegraphics[width=0.42\textwidth]{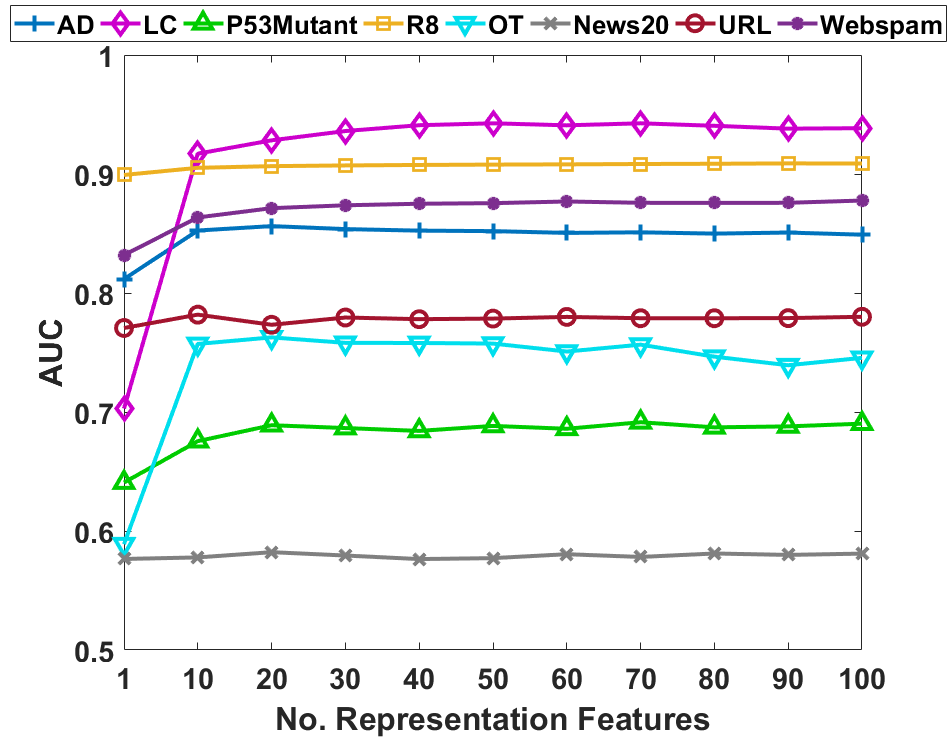}
    \vspace{-2mm}
  \caption{AUC Performance w.r.t. Representation Dimension.}
  \label{fig:sensitivity}
\end{figure}

Figure \ref{fig:sensitivity} presents the AUC performance of REPEN-based Sp using different representation dimensions. REPEN performs very stably when $M$ is set between 10 and 100. Neural network-based methods can represent up to $O(2^{M})$ concepts when using $M$ neural nodes \cite{bengio2013representation}. Therefore, REPEN can theoretically learn complex representations (i.e., up to $O(2^{10})$ concepts) when $M =10$ is used, making it sufficient for most data sets. The expressiveness of REPEN's representations is expected to be increased w.r.t. the representation dimensions. However, the more complex representations we intend to learn, the larger amount of high-quality training data is required. Due to the unsupervised nature of our triplet sampling, our training triplets may not informative enough for REPEN to learn more powerful higher-dimensional (e.g., $M>10$) representations. Hence, the performance of REPEN flattens after $M=10$ (this issue may be addressed by incorporating labeled outliers to generate high-quality training triplets, as explained in the discussion below). We recommend $M=20$ to attain stable accuracy and at the same time make full use of $k$-d tree indexing for distance computation in the representation space.

It is interesting that REPEN using $M=1$ can perform as well as that using a large $M$ in the data sets \textit{R8}, \textit{News20} and \textit{URL}. This phenomenon may be due to two main reasons. On one hand, in the simplest case that the inliers and the outliers are well separable by one linear/non-linear decision boundary, learning a one-dimensional representation space (i.e., $M=1$) is sufficient, since the problem is similar to a one-class or binary classification problem. On the other hand, the optimization in REPEN may also encounter the aforementioned training data quality problem. Improving the triplet quality, e.g., by incorporating a few labeled outliers, can upgrade the performance of REPEN using a large $M$. For example, without using labeled outliers, REPEN obtains similar performance by using $M=1$ and $M=20$ on \textit{News20} and \textit{URL}; however, when we use $l=80$ labeled outliers to improve the triplet quality and learn \textit{1-D} representations, we only obtain AUC results of 0.7201 and 0.8320 on respective \textit{News20} and \textit{URL}, which substantially underperform the AUC results of 0.7707 and 0.9160 based on \textit{20-D} representations, as shown in Figure \ref{fig:dk}. This again highlights the effectiveness of our method in using the limited number of labeled outliers to improve the triplet quality.

\subsection{Scalability Test}

\subsubsection{Experiment Settings}

We generate synthetic data sets by varying the data size in $[1000, 125000]$ of a 10,000-D data set for the scaleup test w.r.t. data size, and likewise, varying the dimensions w.r.t. a fixed data size (i.e., 10,000) for the test w.r.t. dimension. The execution time below includes both training and testing runtime.

\subsubsection{Findings - REPEN Achieves Linear Time Complexity w.r.t. Both Data Size and Dimensionality}

\begin{figure}[h!]
  \centering
    \includegraphics[width=0.47\textwidth]{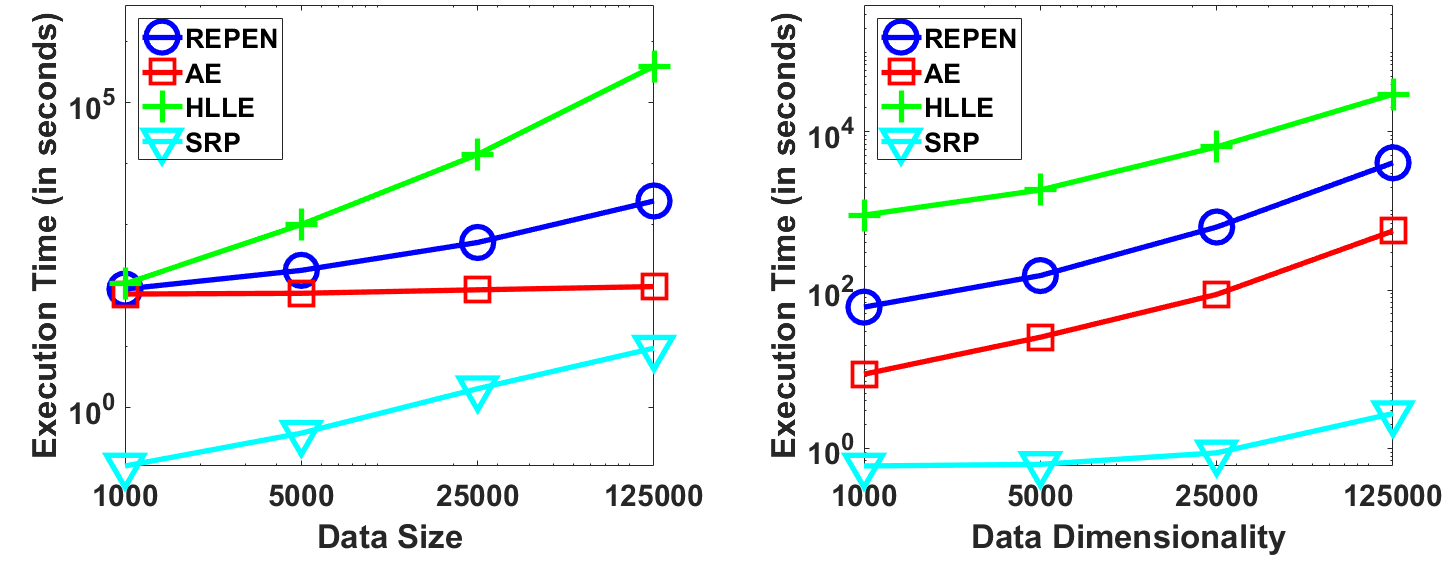}
    \vspace{-2mm}
  \caption{Scalability Test w.r.t. Data Size and Dimensionality. CoP is excluded since it is implemented in a programming language different from the others. Logarithmic scale is used in both axes.}
  \label{fig:scaleup}
\end{figure}

The scalability test results are illustrated in Figure \ref{fig:scaleup}. REPEN has time complexity linear w.r.t. both data size and dimensionality, which justifies our complexity analysis in Section \ref{subsec:algorithm}. REPEN is much faster than HLLE because HLLE requires nearest neighbor searching in the entire data set that has quadratic time complexity w.r.t. data size. SRP is the most efficient method since it only requires the fast operation of random matrix projection. REPEN has a similar network architecture as AE, but it runs slower than AE as it requires more distance computation in its representation learning.

\section{Conclusions}

This paper introduces the RAMODO framework and its instance REPEN that learn customized low-dimensional representations of ultrahigh-dimensional data for random distance-based detectors. By unifying the two correlated tasks, representation learning and outlier detection, we gain three main benefits. One benefit is that we can produce more optimal and stable representations for a specific outlier detector. This is verified by substantially better and more stable detection performance of REPEN, compared to four state-of-the-art representation learning competitors that separate these two correlated tasks. The second benefit is that our method can easily and effectively incorporate application-specific knowledge to learn more application-relevant representations for the given outlier detector, which helps REPEN achieve up to 32\% AUC improvement. The third benefit is that we can effectively represent ultrahigh-dimensional data using very low-dimensional representations (i.e., 20-dimensional representations) and our representations perform stably w.r.t. a wide range of representation dimensions. 

To effectively learn customized representations for the detector, it is critical to generate high-quality training triplets. We show that (i) the combination of state-of-the-art outlier detectors and \textit{Cantelli}'s inequality can help generate sufficiently reliable triplets to learn expressive representations; and (ii) the availability of a few labeled outliers can be further leveraged to substantially improve the triplet quality and learn better representations. 

We are implementing other instances of RAMODO that define sophisticated distance-based outlier scoring to better represent more complex data.

\section*{Acknowledgements}
This work is partially supported by the ARC Discovery Grant DP180100966.

\bibliographystyle{ACM-Reference-Format}
\balance
\bibliography{OutlierDetection}

\end{document}